\documentclass[twoside,11pt]{article}

\usepackage{blindtext}

\usepackage[preprint]{jmlr2e} %

\usepackage{xcolor} %
\colorlet{mygreen}{green!55!black}
\definecolor{myyellow1}{HTML}{D89A3C}
\colorlet{myyellow}{myyellow1!90!black}
\colorlet{myblue}{blue!60!green}
\colorlet{myblue2}{myblue!80!black}
\colorlet{myred}{red!70!black}

\newcommand{\mred}[1]{{\color{myred} #1}}
\definecolor{label1}{HTML}{99292A}
\definecolor{label2}{HTML}{D89A3C}
\definecolor{label3}{HTML}{417481}
\colorlet{label22}{label2!80!black}
\newcommand{\markaa}[1]{\ignorespaces{\color{label1} #1}}
\newcommand{\markbb}[1]{\ignorespaces{\color{label22} #1}}

\usepackage{enumitem}

\usepackage{amsmath}
\allowdisplaybreaks

\usepackage{mathtools}

\usepackage{empheq}

\usepackage{cleveref}
\creflabelformat{equation}{#2\textup{#1}#3}

\usepackage{bbm}

\usepackage{xspace}
\newcommand*{\eg}{{\it e.g.,}\@\xspace}
\newcommand*{\ie}{{\it i.e.,}\@\xspace}

\usepackage{hyperref} %
\hypersetup{
    colorlinks,
    linkcolor={myblue},
    citecolor={myblue},
    urlcolor={myblue}
}

\newcommand\numberthis{\addtocounter{equation}{1}\tag{\theequation}}
    {%
        \end{gathered}\end{equation}
    }

\newcommand{\fracpartial}[2]{\frac{\partial #1}{\partial  #2}}

\newcommand{\br}[1]{\left[#1\right]}
\newcommand{\pr}[1]{\left(#1\right)}

\newcommand{\norm}[1]{\|#1\|}

\newcommand{\T}{\top}

\newcommand{\dt}{\rd t}
\newcommand{\ds}{\rd s}

\def\1{\bm{1}}

\def\rd{{\textnormal{d}}}

\DeclareMathAlphabet{\mathsfit}{\encodingdefault}{\sfdefault}{m}{sl}
\SetMathAlphabet{\mathsfit}{bold}{\encodingdefault}{\sfdefault}{bx}{n}

\def\calL{{\mathcal{L}}}

\def\calN{{\mathcal{N}}}

\def\calP{{\mathcal{P}}}

\def\sR{{\mathbb{R}}}

\newcommand{\E}{\mathbb{E}}

\newcommand{\R}{\mathbb{R}}

\newcommand{\KL}{D_{\mathrm{KL}}}

\DeclareMathOperator*{\argmin}{arg\,min}

\DeclareMathOperator{\Tr}{Tr}

\newcommand*\Zhat{ \widehat{Z} }%
\newcommand*\Yhat{ \widehat{Y} }%
\newcommand*\Psihat{ \widehat{\Psi} }%
\newcommand*\Xbar{ \bar{X} }%
\newcommand*\thetaa{ {\theta^\star} }%
\newcommand*\phii{ {\phi^\star} }%

\usepackage{lastpage}
\jmlrheading{23}{2024}{1-\pageref{LastPage}}{1/21; Revised 5/22}{9/22}{21-0000}{%
    Guan-Horng Liu, Tianrong Chen, Evangelos Theodorou. $^*$Corresponding author. $^\dagger$Work was done while Guan and Tianrong were at Georgia Tech%
}

\ShortHeadings{Deep Generalized Schr{\"o}dinger Bridges}{Liu, Chen, and Theodorou}
\firstpageno{1}

\begin{document}

\title{Deep Generalized Schr{\"o}dinger Bridges: \\ From
    Image Generation to Solving Mean-Field Games}

\author{\name Guan-Horng Liu$^{*,\dagger}$ \email ghliu@meta.com \\
       \addr Fundamental AI Research (FAIR), Meta
       \AND
       \name Tianrong Chen$^{\dagger}$ \email tchen54@apple.com \\
       \addr Apple Machine Learning Research
       \AND
       \name Evangelos A. Theodorou \email evangelos.theodorou@gatech.edu \\
       \addr Georgia Institute of Technology
       }

\editor{My editor}

\maketitle

\begin{abstract}%
    Generalized Schr{\"o}dinger Bridges (GSBs) are a fundamental mathematical framework used to analyze
    the most likely particle evolution based on the principle of least action including kinetic and potential energy.
    In parallel to their well-established presence in the theoretical realms of quantum mechanics and optimal transport, this paper focuses on an algorithmic perspective, aiming to enhance practical usage.
    Our motivated observation is that transportation problems with the optimality structures delineated by GSBs are pervasive across various scientific domains, such as generative modeling in machine learning, mean-field games in  stochastic control, and more.
    Exploring the intrinsic connection between the mathematical modeling of GSBs and the modern algorithmic characterization therefore presents a crucial, yet untapped, avenue.
    In this paper, we reinterpret GSBs as probabilistic models and demonstrate that, with a delicate mathematical tool known as the nonlinear Feynman-Kac lemma, rich algorithmic concepts, such as likelihoods, variational gaps, and temporal differences, emerge naturally from the optimality structures of GSBs.
    The resulting computational framework, driven by deep learning and neural networks, operates in a fully continuous state space (\ie mesh-free) and satisfies distribution constraints, setting it apart from prior numerical solvers relying on spatial discretization or constraint relaxation.
    We demonstrate the efficacy of our method in generative modeling and mean-field games, highlighting its transformative applications at the intersection of mathematical modeling, stochastic process, control, and machine learning.
\end{abstract}


\section{Introduction}

The Schr{\"o}dinger Bridge (SB) problems \citep{schr1931uber,schrodinger1932theorie}, originally proposed by Ewrin Schr{\"o}dinger in the 1930s, ask the following question: What is the most likely probability distribution for i.i.d. Brownian particles at an intermediate time given their distributions observed at two distinct time instants?
Initially introduced to interpret quantum mechanics \citep{fortet1940resolution}, the problem has expanded in scope and gained broader relevance in connection to stochastic control \citep{pavon1991free,dai1991stochastic} and optimal transport \citep{leonard2012schrodinger,chen2021stochastic}.
Nowadays, SBs are well characterized as entropy-regularized optimal transport problems \citep{peyre2017computational,leonard2013survey}, seeking optimal stochastic processes between two distributions that minimize kinetic energy.

Mathematically, given a pair of boundary distributions $\mu(\cdot),\nu(\cdot) \in \calP(\R^d)$,
where $\calP(\R^d)$ is the space of probability distributions,
SB seeks a pair of differential stochastic processes (SDEs) transporting particles between $\mu(\cdot)$ and $\nu(\cdot)$
\begin{subequations}
    \label{eq:sb-sde}
    \begin{align}
        \rd X_t &= \sigma^2\nabla \log \Psi(X_t, t) \dt + \sigma \rd W_t,\qquad X_0 \sim \mu, \quad X_1 \sim \nu, \label{eq:sde}
        \\
        \rd \Xbar_s &= \sigma^2\nabla \log \Psihat(\Xbar_s, s) \ds + \sigma\rd W_s, \qquad \Xbar_0 \sim \nu, \quad \Xbar_1 \sim \mu, \label{eq:rsde}
    \end{align}
\end{subequations}
where $W_{\cdot} \in \R^d$ is the Wiener process, $\sigma\in\R$ is some diffusion scalar known in prior, and $\nabla$ is the gradient operator taken w.r.t. the spatial variable $X_t \in \R^d$.
One can understand $X_t$ as a standard stochastic process evolving \emph{forwardly} along the time coordinate $t\in[0,1]$, whereas $\Xbar_s$ evolves along the \emph{backward} time coordinate $s \coloneqq 1-t$.
These two SDEs are the ``reversed'' process to each other---in the sense that
the path measure induced by \Cref{eq:sde} is equal almost surely to the one induced by \Cref{eq:rsde}.
Furthermore, the functions $\Psi, \Psihat \in C^{2,1}(\R^d, [0,1])$ obey the following optimality conditions characterized by two partial differential equations (PDEs)
\begin{align*}
    \partial_t \Psi(x,t)    = - \frac{1}{2} \sigma^2 \Delta \Psi(x,t), \quad
    &\partial_t \Psihat(x,t) = \frac{1}{2} \sigma^2 \Delta \Psihat(x,t) \\
    \text{s.t. } \Psi(x,0) \Psihat(x,0) = \mu(x), \quad &\Psi(x,1) \Psihat(x,1) = \nu(x),
\end{align*}
where $\Delta$ is the Laplacian operator taken w.r.t. the spatial variable.

\begin{figure}[!t]
    \centering
    \includegraphics[width=.8\linewidth]{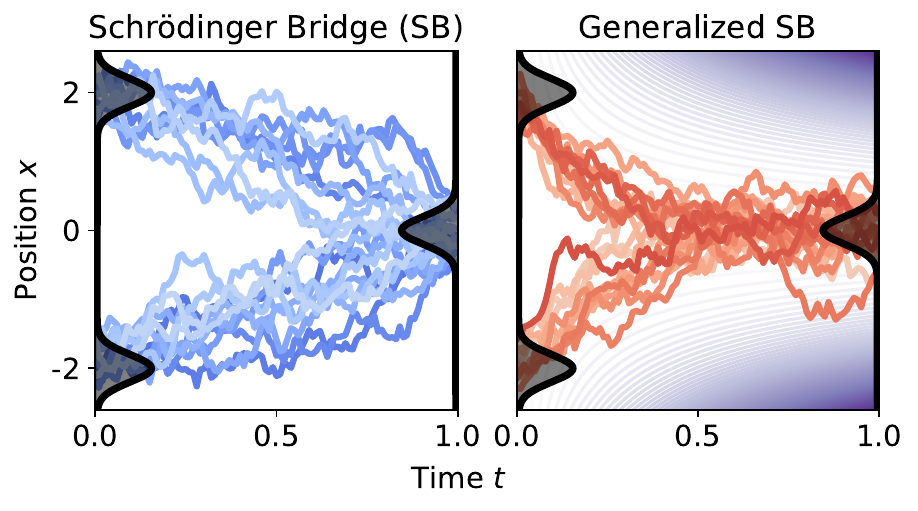}
    \definecolor{sbblue}{RGB}{85,114,235}
    \caption{
        Illustration of \textcolor{sbblue}{Schr\"odinger Bridge (SB)} and \mred{Generalized SB} on an 1-dimensional example w.r.t. a time-varying potential energy (purple contours).
        While SB seeks stochastic processes with minimal kinetic energy, GSB instead considers processes that jointly minimizes both kinetic and potential energies.
    }
    \label{fig:demo}
\end{figure}

Despite that stochastic processes with minimal kinetic energy has proven versatile, finding popularity in signal processing \citep{santambrogio2015optimal,kolouri2017optimal}, computer graphics \citep{solomon2015convolutional,solomon2018optimal}, and statistical inference \citep{cuturi2014fast,srivastava2018scalable},
it corresponds to the squared-Euclidean transport cost among a range of available alternatives \citep{di2017optimal,koshizuka2023neural,cuturi2023monge}.
A crucial extension, as focused in this paper, is the Generalized Schr{\"o}dinger Bridge (GSB), which
introduces an additional layer of complexity by incorporating potential energy into the principle of least action.
This departure from traditional SB models allows for application-specific adaptability, unlocking prevalent problems in general scientific domains where the influence of potential energy is significant, including
economics \citep{achdou2014partial,achdou2022income},
opinion modeling \citep{schweighofer2020agent,gaitonde2021polarization},
population modeling \citep{liu2018mean,achdou2020mean},
computational chemistry and biology \citep{philippidis1979quantum,bunne2023learning}.

The solutions to GSBs obey the same stochastic dynamics as in \Cref{eq:sb-sde}, except that $\Psi, \Psihat$ now solve the following PDEs
\begin{align}
        \partial_t \Psi(x,t)    = V(x,t)\Psi(x,t) - \frac{1}{2} \sigma^2 \Delta \Psi(x,t),& \quad
        \partial_t \Psihat(x,t) = - V(x,t)\Psihat(x,t) + \frac{1}{2} \sigma^2 \Delta \Psihat(x,t)
    \nonumber
    \\
    \text{s.t. } \Psi(x,0) \Psihat(x,0) = \mu(x),& \quad \Psi(x,1) \Psihat(x,1) = \nu(x), \label{eq:sb-pde}
\end{align}
where $V(x,t): \R^d\times[0,1]\rightarrow\R$ is the \emph{potential energy}.
When $V \coloneqq 0$ vanishes, \Cref{eq:sb-pde} reduces to standard SB problems.
In other words, GSBs
\emph{generalizes} SBs with nontrivial potential energy $V(x,t)$.

Due to the coupling at the boundary conditions w.r.t. $\mu$ and $\nu$, the PDEs in \Cref{eq:sb-pde} lack analytic solutions, even in the absence of $V(x,t)$.
Consequently, conventional numerical methods for solving GSBs (and SBs) typically sidestep direct PDE solutions, opting instead for algorithms such as the Sinkhorn methods \citep{sinkhorn1967concerning,knight2008sinkhorn,cuturi2013sinkhorn}, which, despite their superior convergence and stability, require mesh-based discretization \citep{caluya2021wasserstein,chen2023density} of the state space.
Even with the aid of deep learning, modern {deep} PDE solvers struggle to handle the boundary constraints, often necessitating the relaxation of these distributional constraints as soft penalties \citep{ruthotto2020machine,lin2021alternating}.
Given that problems involving the transportation of samples between two distributions are prevalent across scientific applications, with generative modeling \citep{song2021score,lipman2023flow} being notable examples,\footnote{
    In this context, $\mu$ and $\nu$ refer to Gaussian prior and empirical data distribution.
}
the introduction of modern algorithmic concepts in solving GSBs, or how GSBs shall be interpreted from a practical machine learner perspective, remains a crucial yet unexplored avenue.

In this paper, we elucidate the intricate process of deconstructing GSBs into fundamental components, leveraging a delicate mathematical tool known as the nonlinear Feynman-Kac lemma in stochastic process and control \citep{ma1999forward,han2018solving,exarchos2018stochastic}.
From which, we construct a deep learning-based computational framework capable of solving GSBs.
Specifically, we interpret GSBs as probabilistic models by casting them as Neural SDEs \citep{li2020scalable,kidger2021neural}, a family of parametric SDEs wherein drifts (and sometimes diffusions) are parametrized by deep neural networks.
We demonstrate that maximizing their likelihoods effectively enforces the distributional boundary constraints in \Cref{eq:sb-sde}.
Simultaneously, temporal differences, originally appearing in reinforcement learning \citep{wiering2012reinforcement,szepesvari2022algorithms}, play a crucial role in ensuring that particle trajectories adhere to the principle of least action, encompassing both kinetic and, if present, potential energy.
The emergence of these common learning objectives from the PDEs in \Cref{eq:sb-pde} strengthens the intrinsic connection between mathematical modeling, stochastic control, and machine learning.

Built upon these novel learning objectives, we present two practical algorithms, one applicable to solving GSB problems while another specialized for solving SBs, \ie when the primary focus is on determining transportation with minimal kinetic energy.
Given that the latter learning scheme are relatively well-studied, owing to recent advances in training SBs \citep{de2021diffusion,vargas2021solving} despite arising from different perspectives, we provide new theoretical results for the first, more general learning algorithm for GSBs.
Specifically, we prove that jointly optimizing the combined objectives of likelihoods and temporal differences is sufficient for the parametrized Neural SDEs to recover the solutions to GSB.
Highlighting the efficacy and broad applicability of these two learning algorithms, we demonstrate their performance in image generation and solving mean-field games, where the potential energy plays a crucial role in quantifying physical constraints and interactions within individual and large populations.

\section{Methodology}

In this section, we present our deep learning-based computational framework for solving GSBs (and SBs).
The proofs of all theorems can be found in \Cref{app:theorem}.

\subsection{GSBs as Probabilistic Models}

The solutions to the SDEs in \Cref{eq:sb-sde} define forward and backward transport maps between the distributions $\mu$ and $\nu$.
Since the transport maps are, by construction, stochastic, they can be interpreted as probabilistic models---in the sense that one can \emph{generate} samples from, \eg $\nu(\cdot)$ by solving \Cref{eq:sde}, and vise versa.
In this vein, we consider the following representations of GSBs, instantiating a recent class of neural differential equations \citep{chen2018neural} known as Neural SDEs \citep{li2020scalable}
\begin{align}
    \rd X_t^\theta &= \sigma Z^\theta(X_t^\theta, t) \dt + \sigma \rd W_t,\quad~ X_0 \sim \mu, \label{eq:sdeq1} \\
    \rd \Xbar_s^\phi &= \sigma \Zhat^\phi(\Xbar_s^\phi, s) \ds + \sigma \rd W_s,\quad \Xbar_0 \sim \nu, \label{eq:sdeq2}
\end{align}
where $Z^\theta \approx \sigma \nabla \log \Psi$ and $\Zhat^\phi \approx \sigma \nabla \log \Psihat$ are learnable functions, aiming to approximate the drifts in \Cref{eq:sb-sde}, with $\theta$ and $\phi$ gathering their learnable parameters.
The superscripts in $X_t^\theta$ and $\Xbar_s^\phi$ highlight their dependency on the parameters $\theta$ and $\phi$.

\begin{figure}[!t]
    \centering
    \includegraphics[width=.9\linewidth]{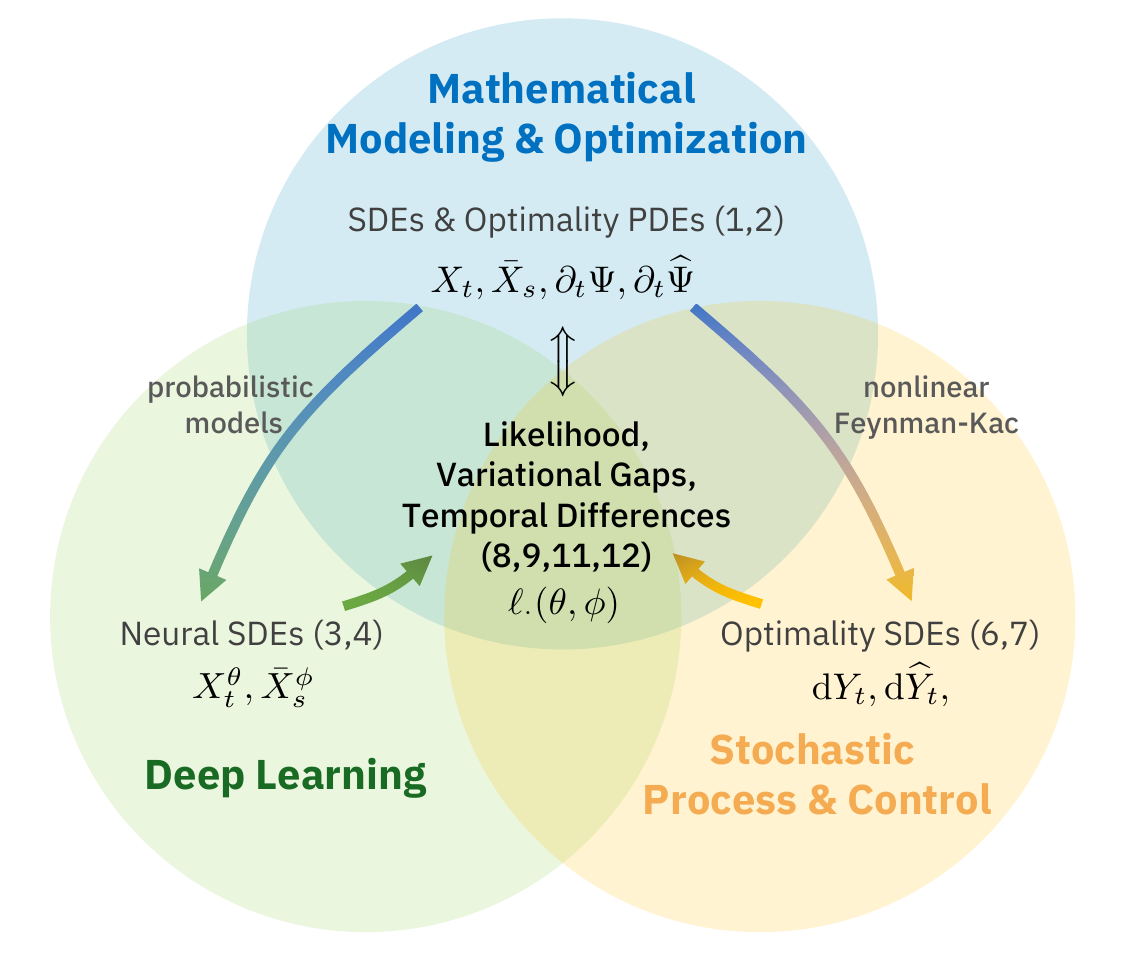}
    \caption{
        Overview of our methodology.
        We begin with the mathematical foundation of Generalized Schr{\"o}dinger Bridges (GSBs) and SBs,
        which are characterized by the SDEs in \Cref{eq:sb-sde} and their optimality conditions represented as PDEs in \Cref{eq:sb-pde}.
        We propose a novel perspective, casting these SDEs as probabilistic models that construct stochastic transport maps between samples drawn from two different distributions, leading to a parametrization with Neural SDEs.
        On the other hand, the optimality PDEs conditions in \Cref{eq:sb-pde} admit stochastic representation and can be transformed into computationally more tractable \emph{optimality SDEs} in \Cref{eq:y,eq:yhat} via a mathematical tool known as the nonlinear Feynman-Kac lemma (\ie \Cref{thm:1}).
        This transformation facilitates the emergence of common learning objectives such as likelihoods in \Cref{eq:nll,eq:nll2}, variational gaps (\Cref{thm:2}), and temporal differences in \Cref{eq:td,eq:td2}.
        We show in \Cref{thm:3} that optimizing the combined objectives provides necessary and sufficient conditions to GSB solutions, making them apt for learning approximate solutions of GSBs.
        In essence, our work reveals an intrinsic connection between mathematical modeling of GSBs, stochastic processes, control, and deep learning.
        }
    \label{fig:method}
\end{figure}

\subsection{Stochastic Representation of \Cref{eq:sb-pde}}

Given trajectories simulated from \Cref{eq:sdeq1}, or \Cref{eq:sdeq2},
we wish to design learning objectives that enforce the optimality conditions in \Cref{eq:sb-pde}.
Unfortunately, the PDEs in \Cref{eq:sb-pde} are not immediately applicable, as they describe the spatial-temporal relation in a point-wise fashion across the product space of $\R^d \times [0,1]$,
as opposed to how the functions shall be evaluated along stochastic processes, which are no where differentiable.
This necessitates a \emph{stochastic} representation of \Cref{eq:sb-pde} in which $(x,t)$ evolves according to some stochastic process, as shown in the following \Cref{thm:1}.
\begin{theorem} \label{thm:1}
    Consider the random variables
    \begin{align}
        \label{eq:nkc}
            Y_t \coloneqq \log \Psi(X_t,t), \quad \Yhat_t \coloneqq \log \Psihat(X_t,t),
    \end{align}
    where $X_t$ solves \Cref{eq:sde}. Then, $Y_t$ and $\Yhat_t$ solve
        \begin{align}
            &\rd Y_t = \pr{\frac{1}{2} \norm{\sigma \nabla Y_t}^2 + V } \dt + \sigma \nabla Y_t \cdot \rd W_t, \label{eq:y} \\
            &\rd \Yhat_t = \pr{\frac{1}{2} \norm{\sigma \nabla \Yhat_t}^2 - V +
                            \nabla \cdot (\sigma^2 \nabla \Yhat_t) + \sigma^2 \nabla \Yhat_t \cdot \nabla Y_t}
                        \dt + \sigma \nabla \Yhat_t \cdot \rd W_t.
            \label{eq:yhat}
        \end{align}
        The solutions to the PDEs in \Cref{eq:sb-pde} can be recovered via taking conditional expectation:
        \begin{align*}
            \E[Y_t | X_t=x] = \log \Psi(x,t), \quad \E[\Yhat_t | X_t=x] = \log \Psihat(x,t).
        \end{align*}
\end{theorem}
Theorem~\ref{thm:1} stems directly from the nonlinear Feynman-Kac lemma \citep{ma1999forward,han2018solving}, providing a stochastic representation (in terms of $Y_t$ and $\Yhat_t$) of the optimality PDE conditions in \Cref{eq:sb-pde} by characterizing how their function values change along the SDE $X_t$ in \Cref{eq:sde}.
Notably, their values evolve as another set of SDEs.
Such transformations, bridging specific classes of PDEs and SDEs, have become prominent in recent advancements in
numerical methods for stochastic control \citep{exarchos2018stochastic,pereira2019learning}.
Importantly, rather than solving the PDEs in the entire function space,
it suffices to solve them locally around high probability regions induced by the SDE.
We will solidify the connection to stochastic control shortly.

\subsection{Likelihood Objective}

With Theorem~\ref{thm:1}, we can now derive the model likelihood \citep{myung2003tutorial}, a prevalent objective in learning probabilistic models.
The standard procedure is to minimize the negative log-likelihood, which, in our case, would be computed from the Neural SDEs.
Consider the \emph{forward} Neural SDE in \Cref{eq:sdeq1}, for instance.
Given a trajectory simulated from $X_0\sim \mu$,
our goal is to estimate
\begin{align*}
    \log\mu^\theta(X_0) \approx \log\Psi(X_0,0) + \log\Psihat(X_0,0),
\end{align*}
where the approximation follows from the boundary condition in \Cref{eq:sb-pde}.
In other words, we wish to estimate $\log\Psi$ and $\log\Psihat$ given $X_{t\in[0,1]}^\theta$. Applying Theorem~\ref{thm:1} suggests the objective
\begin{align}
    \label{eq:nll}
    \ell_\mathrm{fwd}(\theta, \phi) = -\E \br{Y_0 + \Yhat_0 | X_0 } = \int_0^1 \E\br{ \frac{1}{2}\norm{Z_t^\theta {+} \Zhat_t^\phi} + \nabla \cdot (\sigma \Zhat_t^\phi) } \dt - \E \log \nu(X_1^\theta),
\end{align}
where we substitute $\sigma\nabla Y_t \approx Z^\theta(X_t^\theta,t)$, $\sigma\nabla \Yhat_t \approx \Zhat^\phi(X_t^\theta,t)$ and shorthand them respectively by $Z_t^\theta$ and $\Zhat_t^\phi$.
Notice that the It{\^o} integrals vanish due to the martingale property.

Repeating similar derivation for the backward Neural SDE in \Cref{eq:sdeq2}, except now along the reversed time coordinate $s$, yields the likelihood objective for $\log \nu^\phi(\Xbar_0)$
\begin{align}
        \ell_\mathrm{bwd}(\phi, \theta) = \int_0^1 \E\ \left[ \frac{1}{2}\norm{Z_s^\theta {+} \Zhat_s^\phi} + \nabla \cdot (\sigma Z_s^\theta) \right] \ds - \E\log \mu(\Xbar_1^\phi),
        \label{eq:nll2}
\end{align}
where we substitute $\sigma\nabla Y_s \approx Z^\theta(\Xbar_s^\phi,s) \equiv Z_s^\theta$ and $\sigma\nabla \Yhat_s \approx \Zhat^\phi(\Xbar_s^\theta,s) \equiv \Zhat_s^\phi$, similar to \Cref{eq:nll}.

\subsection{Variational Gap of $\ell_\text{fwd}$ and $\ell_\text{bwd}$}

It should be noted that \Cref{thm:1} provides characterization only at the equilibrium, \ie when $X_t$'s follow the \emph{optimal} solutions in \Cref{eq:sde}.
Due to the parametrization, \Cref{eq:nll,eq:nll2} present the variational bounds.
This raises questions of
whether minimizing \Cref{eq:nll,eq:nll2} suffices to close the variational gaps.
We provide a positive answer in the following result.
\begin{theorem}\label{thm:2}
    Let $p^\theta$ and $p^\phi$ denote the path densities induced respectively by the Neural SDEs in \Cref{eq:sdeq1,eq:sdeq2}. Then,
    \begin{align*}
        \ell_\mathrm{fwd}(\theta, \phi) = \KL(p^\theta || p^\phi) - \log \mu^\theta(X_0), \quad
        \ell_\mathrm{bwd}(\phi, \theta) = \KL(p^\phi || p^\theta) - \log \nu^\phi(\Xbar_0), %
    \end{align*}
    where $\KL(\cdot || \cdot)$ is the Kullback-Leibler (KL) divergence, and $\mu^\theta$ and $\nu^\phi$ are the parametrized model likelihoods. Furthermore, their optimal values achieve the true likelihoods, \ie
    \begin{align*}
        -\log \mu(X_0) = \min_{\theta,\phi} \ell_\mathrm{fwd}(\theta,\phi), ~~
        -\log \nu(\Xbar_0) = \min_{\phi,\theta}\ell_\mathrm{bwd}(\phi,\theta).
    \end{align*}
\end{theorem}
Theorem \ref{thm:2} suggests that the variational gaps can be quantified by the KL divergences between two Neural SDEs.
Take $\ell_\mathrm{fwd}(\theta, \phi)$ for instance.
Since $\phi$ appears only in the KL divergence, the minimizer $\phi$ must obey $p^{\phi^\star} {=}~ p^\theta$ for $t\in [0,1)$.
Minimizing $\ell_\mathrm{fwd}(\theta, \phi)$ w.r.t. $\theta$ thereby amounts to maximum likelihood estimation, justifying the use of \Cref{eq:nll,eq:nll2} in practice.
Essentially,
the likelihood objectives
$\ell_\mathrm{fwd}(\theta,\phi)$ and
$\ell_\mathrm{bwd}(\phi,\theta)$
ensure the terminal distributions of the Neural SDEs in \Cref{eq:sdeq1,eq:sdeq2} to satisfy the distributional constraints.
This, crucially, distinguishes our proposed framework from prior attempts built also upon neural differential equations \citep{ruthotto2020machine,lin2021alternating}, which require \emph{relaxing} the distributional constraint with soft penalty.
We highlight this distinction arising from the delicate transformation of the PDE constraints in \Cref{eq:sb-pde} to the SDEs in \Cref{eq:y,eq:yhat}.

Despite being encouraging, the results in Theorem \ref{thm:2} is \emph{insufficient} in showing the minimizers of $\ell_\mathrm{fwd}$, or $\ell_\mathrm{bwd}$, solve GSBs.
This is best explained by revisiting \Cref{eq:y,eq:yhat}, and
notice that the two terms related to potential energy, $+V$ and $-V$, cancel out in the computation of the likelihood objective in \Cref{eq:nll}.
In other words,
the likelihood objectives
remain the same regardless of the choices of $V(x,t)$,
ensuring the distributional boundary constraints yet not necessarily the optimality of the trajectories in between.
Below, we discuss a simple remedy inspired by stochastic control, further strengthening the {interconnection} of our proposed framework across diverse research fields.

\subsection{Temporal Difference (TD) Objective}

Let us revisit the random variable $Y_t$ and its connection to the PDEs in \Cref{eq:sb-pde}.
These PDEs admit an intriguing variational formulation, in that they are the optimality conditions to
following constrained optimization
\begin{equation}
    \label{eq:sb-soc}
    \begin{split}
    &\min_{u} \int_0^1 \E \br{ \frac{1}{2}\|u(X_t,t)\|^2 + V(X_t,t) } \dt, \quad
    \text{s.t. } \rd X_t = \sigma u(X_t, t) \dt + \sigma \rd W_t,~ X_0 \sim \mu,~ X_1 \sim \nu.
\end{split}
\end{equation}
The variational formulation provides an intuitive interpretation to the potential energy $V(x,t)$.
Essentially, GSB solves a stochastic optimal control (SOC) with distributional boundary constraints, where, among all the controlled processes that transport samples from $\mu(\cdot)$ to $\nu(\cdot)$, it considers the one with the least kinetic, presenting as the velocity norm, and potential energy to be optimal.

Serving as the optimality conditions to the constrained SOC in \Cref{eq:sb-soc}, the PDEs in \Cref{eq:sb-pde} share the same roles as the Hamilton-Jacobi-Bellman (HJB) equation appearing in SOC literature \citep{kappen2005path,exarchos2018stochastic}, with $\log \Psi(x,t)$ resembling the value function.\footnote{
    These interpretations can be made concrete; see \Cref{app:b.2} for more detailed discussions.
}
In other words,
the random variable $Y_t \coloneqq \log \Psi(X_t,t)$ can be understood as the unique stochastic representation of the value function evaluated along the process $X_t$.
This implies that the dynamics of $Y_t$ evolve according to the Bellman equation \citep{bellman1954theory}, except in a stochastic, continuous-time fashion.
Indeed, discretizing \Cref{eq:y} with some fixed step size $\delta t$ and $\delta W_t \sim \calN(0, \delta t)$ yields
\begin{align*}
    Y_{t+\delta t} = Y_t +
      \markaa{\pr{\frac{1}{2} \norm{\sigma\nabla Y_t}^2 + {V}}\delta t}
    + \markbb{\sigma\nabla Y_t \cdot \delta W_t},
\end{align*}
which resembles a non-discounted temporal difference \citep{todorov2009efficient,lutter2021value}, with the distinction that, beyond the usual \markaa{``\textit{rewards}'' associated with control and state costs}, a \markbb{stochastic term} is introduced.
Unlike the conventional Bellman equation where this stochastic term disappears upon expectation, here it plays a critical role in characterizing the intrinsic stochasticity of $Y_t$ and exhibits further connections to the optimal control variate \citep{robert1999monte,richter2022robust}.

With the interpretation in mind,
we consider $Y^\theta \approx \log \Psi$ and $\Yhat^\phi \approx \log \Psihat$ and re-parametrize the drifts of Neural SDEs by $Z_t^\theta \coloneqq \sigma\nabla Y^\theta$ and $\Zhat_t^\phi \coloneqq \sigma\nabla \Yhat^\phi$. Then, the TD objectives can be constructed as follows
\begin{align}
    \label{eq:td}
    &\ell_\mathrm{TD}(\theta) = \E \Big\|
        Y_{t{+}\delta t}^\theta {-} \Bigl(
            Y_{t}^{\breve\theta} + \Bigl(
                \frac{1}{2} \norm{Z_t^{\breve\theta}}^2 {+} V
            \Bigr) \delta t + Z_t^{\breve\theta} \cdot \delta W_t
        \Bigr)
    \Big\|^2, \\
    \label{eq:td2}
    &\ell_\mathrm{TD}(\phi) = \E \Big\|
        \Yhat_{t{+}\delta t}^\phi {-} \Bigl(
            \Yhat_{t}^{\breve\phi} + \Bigl(
                \frac{1}{2} \norm{\Zhat_t^{\breve\phi}}^2 {-} V + \nabla \cdot (\sigma \Zhat_t^{\breve\phi}) + \Zhat_t^{\breve\phi} \cdot Z^{\breve\theta}_t
            \Bigr) \delta t + \Zhat_t^{\breve\phi} \cdot \delta W_t
        \Bigr)
    \Big\|^2,
\end{align}
where $\breve\theta$ and $\breve\phi$ denote the detached copy of the parameters.
The expectations in \Cref{eq:td,eq:td2} are taken w.r.t. the stochasticity induced in the Neural SDE in \Cref{eq:sdeq1}, with $t$ sampled uniformly between $[0,1]$.
Our final result suggests that adding these TD objectives to the likelihood objective suffices to recover GSBs.
\begin{theorem} \label{thm:3}
    $Y^\theta$, $\Yhat^\phi$, $Z^\theta$, and $\Zhat^\phi$ solve the GSB problem if and only if they are the minimizers of $\ell_\mathrm{fwd}(\theta,\phi) + \ell_\mathrm{TD}(\phi)$
\end{theorem}
Theorem~\ref{thm:3} can be easily extended to different combinations with $\ell_\mathrm{bwd}(\phi,\theta)$ and $\ell_\mathrm{TD}(\theta)$.

\subsection{Joint and Alternate Training Schemes}

Theorem~\ref{thm:3} validates a training scheme where we jointly learn the parameters $\theta, \phi$
\begin{align}
    \label{eq:joint}
    \min_{\theta, \phi} \ell_\mathrm{fwd}(\theta,\phi) + \ell_\mathrm{TD}(\phi).
\end{align}
Though joint training works well in low-to-moderate dimensional problems, it necessitates back-propagating through the Neural SDE (notice, for instance, that computing $\ell_\mathrm{fwd}(\theta,\phi)$ requires simulating $X_{t\in[0,1]}^\theta$), which becomes prohibitively expansive as the dimension $d$ grows.
An alternative is to alternate between minimizing $\ell_\mathrm{fwd}(\theta,\phi)$ and $\ell_\mathrm{fwd}(\phi,\theta)$ but only w.r.t. the second input:
\begin{equation}
    \label{eq:alt}
    \begin{split}
        \phi^{(k)}   \leftarrow& \argmin_{\phi} \ell_\mathrm{fwd}(\theta^{(k)},\phi), \qquad
        \theta^{(k+1)} \leftarrow \argmin_{\theta} \ell_\mathrm{bwd}(\phi^{(k)},\theta).
    \end{split}
\end{equation}
From \Cref{thm:2}, \Cref{eq:alt} is equivalent to alternate minimization of the KL divergences, as the likelihoods are now independent of the optimized variables and hence can be dropped.
This alternate procedures instantiates the Iterative Proportional Fitting algorithm \citep{kullback1968probability}, commonly adopted for solving SBs \citep{,vargas2021solving}.
Despite being applicable only when $V \coloneqq 0$, it stands as a pragmatic adjustment when least kinetic energy alone suits the applications, as demonstrated in the subsequent section.

\section{Numerical Experiments}

In this section, we validate our proposed deep learning-based framework, \ie \Cref{eq:joint,eq:alt}, in solving SBs and GSBs. We focus on two distinct applications---generative modeling and solving mean-field games---aiming to highlight the extensive applicability of SBs and GSBs in diverse scientific areas.

\subsection{Generative Modeling with Alternate Training}

Generative modeling stands at the forefront of modern machine learning, playing a pivotal role in transforming various industries and everyday experiences with cutting-edge tools like GPT-4 \citep{achiam2023gpt} and Stable Diffusion \citep{rombach2022high}.
At its core, generative models aim to discern the underlying data distribution given only empirical samples. Once learned, the models can be used to sample or, equivalently, \emph{generate} new data and content.
Advanced deep generative models, exemplified by flow or diffusion models \citep{lipman2023flow,song2021score}, strive to identify neural differential equations capable of mapping simple distributions, typically Gaussian, into complex, often intractable, data distributions. Adopting our notations, let
\begin{align*}
    \mu(x) \coloneqq \calN(x), \qquad \nu(x) \coloneqq p_\text{data}(x),
\end{align*}
where $\calN$ and $p_\text{data}$ denotes respectively normal and data distributions.
Then, SBs solve a set of Neural SDEs transporting samples back-and-forth between the two distributions with minimal kinetic energy. Notably, the kinetic regularization has been shown effectively in enhancing training and sampling efficiency \citep{finlay2020train,chen2022likelihood,shaul2023kinetic}.

Regarding the choice of $p_\text{data}$, we explore three data sets, namely {MNIST} \citep{lecun2010mnist}, {CelebA} \citep{liu2015faceattributes}, and {Cifar10} \citep{Krizhevsky09learningmultiple}, each representing different image domains.
MNIST consists of 60,000 handwritten digits, CelebA features 202,599 images of celebrities' faces, and Cifar10 contains 60,000 objects categorized into ten classes, such as airplanes, cars, birds, cats, deer, dogs, frogs, horses, ships, and trucks.
These data sets are selected not only for their popularity in the machine learning community but to diversify the distribution being modeled.
MNIST represents discrete data in gray scale, while both CelebA and Cifar10 consist of color images.
Additionally, Cifar10 encompasses a variety of natural images across different categories, whereas CelebA focuses on facial attributes.
All images are resized to 32 by 32 resolution, resulting in problem dimensions of $d=1024$ for MNIST and $d=3072$ for CelebA and Cifar10.
We employ the alternate training scheme in \Cref{eq:alt} to train an SB on each data set.

\begin{figure*}[bt!]
    \includegraphics[width=\linewidth]{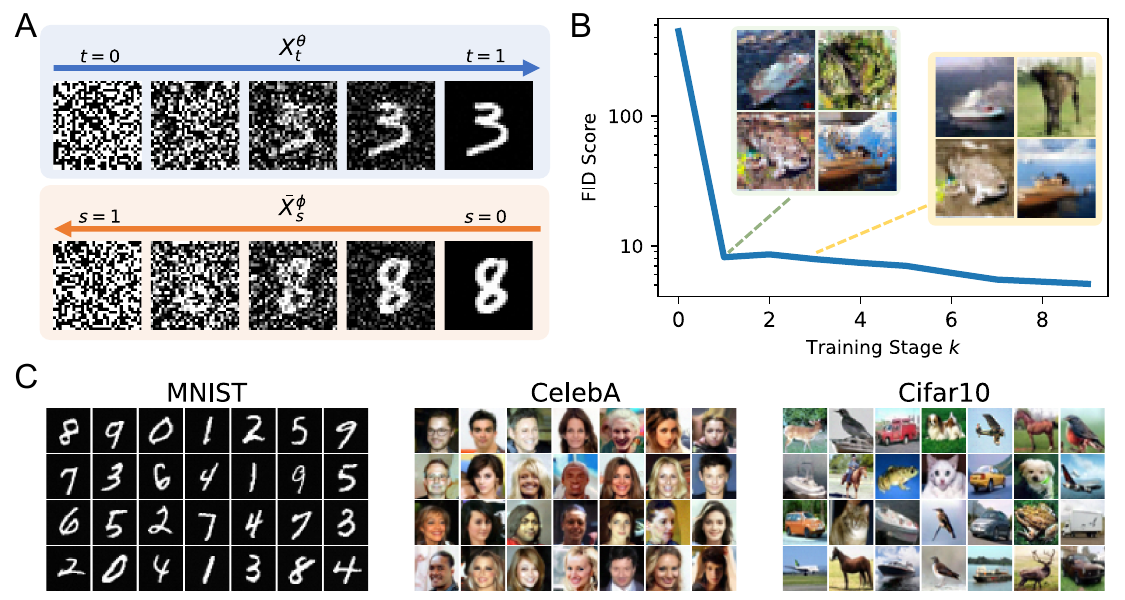}
    \caption{
        Results of generative modeling in image domains.
        \textit{(A)} Simulations of the learned Neural SDEs, exemplified with a handwritten digit data set MNIST. While $X_t^\theta$ evolves along the forward time coordinate $t$, transforming a Gaussian noise to an image (in this case a digit 3), $\Xbar_s^\phi$ instead moves backward along the coordinate $s \coloneqq 1-t$ and generates random noises.
        \textit{(B)} Training progress of Cifar10 with respect to the Fr\'echet Inception Distance (FID) score. Notice how the the FID score stably decreases, improving the fidelity of the generated images over time.
        \textit{(C)} Generated samples, \ie $X_1^\theta$, on all three tested data sets, including MNIST (left), CelebA (middle), and Cifar10 (right).
    }
    \label{fig:gen}
\end{figure*}

The learning results are summarized in Fig.~\ref{fig:gen}, demonstrating the effectiveness of the proposed computational framework as a generative model in synthesizing high-fidelity images, by learning transport maps between Gaussian priors and image distributions.
Notably, in Fig.~\ref{fig:gen}\textit{A}, the forward Neural SDE $X_t^\theta$ learns to generate images from random noises, while the backward Neural SDE $\Xbar_s^\phi$ reverses the generative process.
This training of SBs is therefore reminiscent of diffusion models, as pointed out in \citet{de2021diffusion,vargas2021solving,chen2022likelihood}, except that the stochastic processes from data to noises are learned rather than prescribed.
Quantitatively, Fig.~\ref{fig:gen}\textit{B} tracks the performance of modeling Cifar10 during training, measured by the Fr\'echet Inception Distance (FID) score \citep{heusel2017gans},
a commonly-used metrics for evaluating the proximity of the generated distribution to the empirical data distribution.
The decreasing FID score throughout training indicates an enhancement in the quality of generated images. This improvement is further confirmed by examining images generated at different training stages while freezing the random seeds and initial states, revealing a distinguishable reduction in visible flaws.

\subsection{Mean-Field Games (MFGs) with Joint Training}

MFGs constitute a crucial mathematical framework in modeling the collective behavior of individual agents interacting stochastically with a large population.
This framework is prevalent across multidisciplinary scientific areas
\citep{achdou2014partial,achdou2022income,schweighofer2020agent,gaitonde2021polarization,liu2018mean,achdou2020mean}, delineating a noncooperative differential game
involving \emph{a continuum population of rational agents}.
Here, ``continuum'' refers to the scenario where the number of agents in a multi-agent game approaches infinity (hence the term \emph{mean-field}), resulting in the interaction between individuals and population density.
On the other hand, ``rationality'' relies on some variational structure to which the collective behavior of individuals must adhere.

We focus on classical examples of crowd navigation in $\R^2$. Our goal is to understand how rational agents navigate between the population distributions observed at two distinct snapshots. Rationality, in this context, involves several objectives:
\begin{enumerate}
    \item Minimizing execution efforts from one point to another.
    \item Considering interactions between individuals and population, such as avoiding crowded or congested regions with high probability density.
    \item Avoiding physical constraints, such as obstacles.
\end{enumerate}
To explore how GSBs can be applied to solve MFGs, we observe that the first objective aligns with the kinetic energy, representing the total control spent in transporting samples between two distributions. Meanwhile, the second and third objectives influence agents' behaviors by imposing preferences through their ``state'' rather than the executed control, making them applicable as the potential energy $V(x,t)$.
Indeed, crowd navigation MFGs are conventionally framed as the distribution-constrained SOC \citep{ruthotto2020machine,lin2021alternating}, where $V(x,t)$ either measures the cost incurred for individuals to stay in densely crowded regions with high population density such as entropy or congestion cost \citep{ruthotto2020machine,lin2021alternating} or restricted area such as obstacles
\begin{align*}
    V_\text{ent}(x,t) \coloneqq  \log \rho(x,t),\quad
    V_\text{cgst}(x,t) \coloneqq  \E_{y\sim \rho}\br{\frac{1}{\norm{x{-}y}^2{+}1}},\quad
    V_\text{obs}(x,t) \coloneqq c \cdot \mathbbm{1}_\text{obs}(x),
\end{align*}
where $c$ is a tuning hyper-parameter and $\mathbbm{1}_\text{obs}(\cdot)$ denotes the indicator of the problem-dependent obstacle set.

\begin{figure*}[bt!]
    \includegraphics[width=\linewidth]{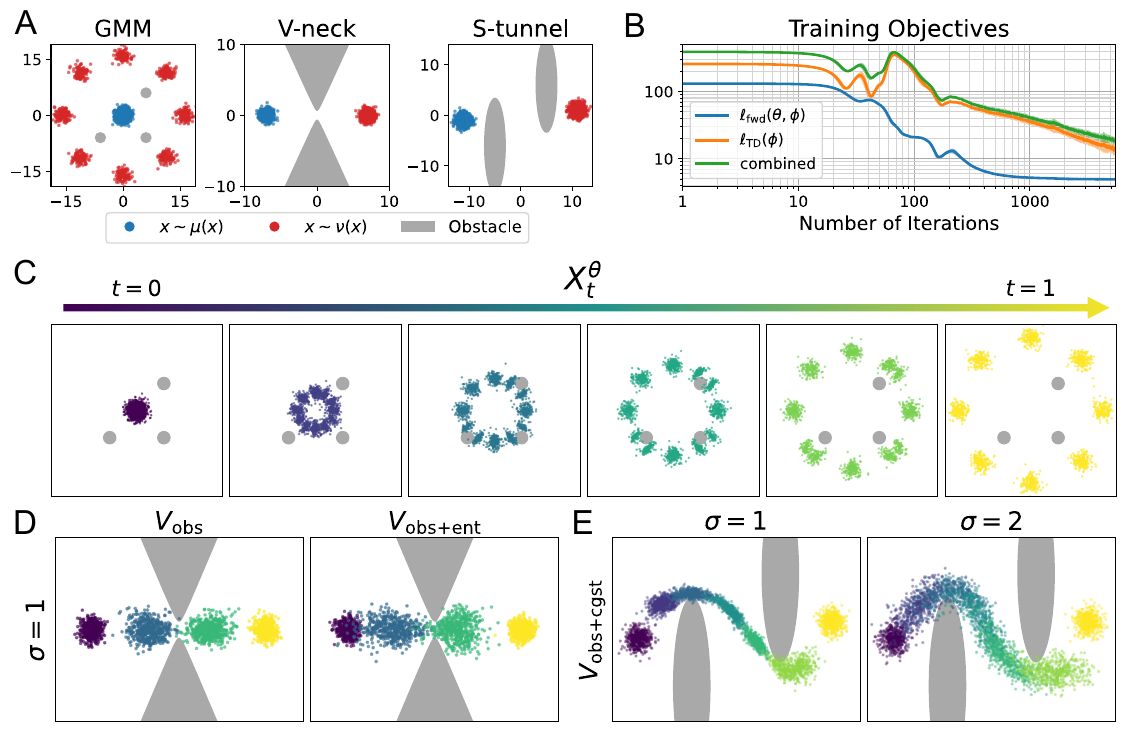}
    \caption{
        Results of mean-field games (MFGs) in crowd navigation scenarios.
        \textit{(A)} The configurations of three distinct MFGs, namely GMM, V-neck, and S-tunnel, showcasing the boundary distributions $\mu$ (blue), $\nu$ (red), and restricted areas (gray).
        \textit{(B)} Training progress of GMM, providing insights into how the likelihood and temporal difference (TD) objectives contribute to the combined training objective. We report the average over 5 random seeds.
        \textit{(C)} The simulation results on GMM, where seven population snapshots are displayed at regular intervals between $t\in[0,1]$, each with a different color.
        \textit{(D and E)} Additionally, an ablation study is conducted on the entropy interaction cost and diffusion coefficient $\sigma$ for V-neck and S-tunnel, respectively. For optimal visualization, four population snapshots are presented for V-neck, and seven for S-tunnel.
    }
    \label{fig:mfg}
\end{figure*}

We examine three crowd navigation MFGs from the literature \citep{ruthotto2020machine,lin2021alternating,liu2022deep}, encompassing
\textit{(i)} asymmetric obstacle avoidance,
\textit{(ii)} entropy interaction with a V-shaped bottleneck,
and \textit{(iii)} congestion interaction on an S-shaped tunnel.
Figure~\ref{fig:mfg}\emph{A} depicts the configuration of each MFG, denoted as \textit{GMM}, \textit{V-neck}, and \textit{S-tunnel}, respectively.
For simplicity, we consider Gaussians or mixtures of Gaussians for $\mu,\nu$, even though our method is applicable to general distributions, as demonstrated in the previous example.
On all three MFGs, we train Generalized Schr{\"o}dinger Bridges with the combined objectives of likelihoods and temporal differences (TDs) in \Cref{eq:joint}.
As illustrated in Fig.~\ref{fig:mfg}\emph{B}, the likelihood objectives typically converge faster than the TD objectives,
and the overall training saturates within 4,000-5,000 iterations.
Empirically, we observe that joint training of $\theta,\phi$ demonstrates faster convergence than its alternate counterpart, albeit with higher memory consumption.

Our primary outcomes are illustrated in Fig.~\ref{fig:mfg}\emph{C-E}.
It is evident that, across all three crowd navigation setups, our method effectively guides the population to navigate smoothly while avoiding restricted areas (marked gray).
A notable difference emerges when the interaction entropy cost is introduced, as shown in Fig.~\ref{fig:mfg}\emph{D}, where the population is encouraged to disperse both before and after navigating through the V-shaped bottleneck.
Meanwhile, Fig.~\ref{fig:mfg}\emph{E} demonstrates the robustness of our method w.r.t. different diffusion coefficients $\sigma$.
Importantly, the learned transport maps satisfy the required distributional boundary constraints in all cases.
In summary, our computational framework, serving as a numerical solver for MFGs, exhibits robust performance across diverse setups involving various optimality structures, degrees of stochasticity, and problem configurations.

\section{Discussion and Outlook}

We present a deep learning-based method for solving Generalized Schr{\"o}dinger Bridges (GSBs), employing a set of Neural SDEs to transport samples between two distributions while adhering to the variational principle of kinetic and potential energy.
Our proposed computational framework falls within the realm of numerical solvers for neural differential equations \citep{chen2018neural}, playing a crucial role across diverse scientific domains and bridging the gap between mathematicians and machine learning practitioners.
Conventional solvers treat deep neural networks merely as function approximations and base their
algorithms on standard optimization procedures, such as Lagrangian \citep{ruthotto2020machine} and PDE regression \citep{raissi2019physics}.
Our work, in its essence, challenges this paradigm by revealing the inherent connection between variational structures in mathematical models and fundamental concepts in machine learning.
This necessitates a fundamental understanding between mathematical modeling, optimization, and deep learning, through which we emphasize the significance of the nonlinear Feynman-Kac lemma, a key tool in the analysis of stochastic processes and control.
The integration of machine learning components results in a computational framework that transcends the application boundaries to which the developed method can be applied.
We demonstrate this advantage through various applications that may seem unrelated at first glance, ranging from generative modeling for image domains to solving crowd navigation mean-field games in computational social science.

It is worth noticing that the TD objectives inherited in the optimality SDEs \Cref{eq:y,eq:yhat} narrow the gap between GSBs and Deep Reinforcement Learning (RL).
Notably, the variational formulation in \Cref{eq:sb-soc} resembles a standard RL problem, seeking an optimal policy that minimizes costs (or equivalently, maximize rewards) along trajectories.
However, GSBs distinguish themselves from standard RL setups by introducing a distributional boundary constraint instead of a terminal cost, addressed through an augmented learning objective with a likelihood component.
The intriguing use of likelihoods to enforce distribution constraints opens the door to potential advancements leveraging algorithmic developments in the RL community.
The application of numerical methods designed for distribution-constrained RL or Stochastic Optimal Control holds promise for providing solutions to previously challenging mean-field games and other social science problems under more realistic settings,
aligning with scenarios where social scientists observe population snapshots over time and aim to understand behavior in between these observed intervals or future evolution.

Looking ahead, there are several open questions that warrant exploration.
From an application standpoint, the extension of the objective in \Cref{eq:sb-soc} to encompass general Lagrangian functions $\calL(x, \dot x, t)$ holds potential significance, as it could unlock applications in metric learning, computational chemistry, and biology. These domains often involve intricate geometry structures that go beyond Euclidean spaces.
Furthermore, the extension to incorporate other kinds of stochastic processes, such as jump diffusions, could open avenues for applications in areas like operations research, time series analysis, and finance.

\acks{   
    We thank Ioannis Exarchos and Oswin So for their generous involvement and helpful supports;
    Marcus A Pereira and Ethan N Evans for their participation in the early stage of project exploration;
    and Bogdon Vlahov for sharing the computational resources.
    This research was supported by the ARO Award  \#W911NF2010151 and DoD Basic Research Office Award HQ00342110002.
}

\newpage

\appendix
\section{Proofs}
\label{app:theorem}

\subsection{Preliminaries}
We first state some useful equalities that will appear in the proceeding proofs.

\begin{itemize}[leftmargin=1.5em]
    \item
    For any point $x\in \R^d$ such that $p(x) \neq 0$, it holds that
    \begin{align*}
        \frac{1}{p(x)}\Delta p(x) &= 
        \frac{1}{p(x)} \nabla \cdot \nabla p(x)
        = \frac{1}{p(x)} \nabla \cdot \Bigl( p(x) \nabla \log p(x) \Bigr) \\
        &= \frac{1}{p(x)} \Bigl( \nabla p(x) \cdot \nabla \log p(x) + p(x) \Delta \log p(x) \Bigr) \\
        &= 
        \norm{\nabla \log p(x)}^2 + \Delta \log p(x)\numberthis \label{lm:1}
    \end{align*}
    \item Under mild assumptions \citep{anderson1982reverse,yong1999stochastic}
    such that all distributions approach zero at a sufficient speed as $\norm{x} \rightarrow \infty$, and that all integrands are bounded, we have
    \begin{align}
        \E_{x\sim p(x)} \br{\Delta \log q(x)} = \E_{x\sim p(x)} \br{\nabla \cdot \nabla\log q(x) } = \E_{x\sim p(x)} \br{ - \nabla \log p(x) \cdot \nabla\log q(x) },
        \label{eq:int-by-part}
    \end{align}
    where the second equality follows by integration by parts and reparameterization trick
    \begin{align*}
        \int p(x) \Bigl( \nabla \cdot \nabla\log q(x)     \Bigr) \rd x
        &= \int    - \Bigl( \nabla p(x) \cdot \nabla\log q(x) \Bigr) \rd x \\
        &= \int - p(x) \Bigl( \nabla \log p(x) \cdot \nabla\log q(x) \Bigr) \rd x.
    \end{align*}
    More generally, under the same regularity, it holds for a vector field $Z: \R^d \mapsto \R^d$ that
    \begin{align}
        \E_{x\sim p(x)} \br{\nabla \cdot Z(x) } = \E_{x \sim p(x)} \br{ - \nabla \log p(x) \cdot Z(x) }.
        \label{eq:int-by-part2}
    \end{align}
    
    \item It{\^o} formula \citep{ito1951stochastic}:
    Let $X_t$ be the solution to the It\^o SDE
    \begin{align*}
    \rd X_t = f(X_t, t) \dt + \sigma \rd W_t.
    \end{align*}
    Then, the stochastic process $v(X_t,t)$, where $v \in C^{2,1}(\sR^d,[0,1])$, is also an It{\^o} process
    \begin{align*}
        \rd v(X_t,t) =
            \fracpartial{v(X_t,t)}{t}\dt
            &+ \br{ \nabla v(X_t, t) \cdot f + \frac{1}{2} \sigma^2 \Delta v(X_t, t)} \dt
            + \sigma \nabla v(X_t, t) \cdot \rd W_t
            \numberthis \label{eq:ito}
    \end{align*}

    \item 
    {Nonlinear Feynman-Kac lemma \citep{}:}
    The lemma establishes an innate connection between different classes of PDEs and
    forward-backward SDEs. Consider the following forward-backward SDEs,
    \begin{subequations}
        \begin{empheq}[left={\empheqlbrace}]{align}
              \rd X_t &=  f(X_t, t) \dt + G(X_t, t) \rd W_t, \quad  X_0 = x_0 \label{eq:fbsde-f} \\
              \rd Y_t &=  - h(X_t, Y_t, Z_t, t) \dt + Z \cdot \rd W_t, \quad Y_1 = \varphi(X_1) \label{eq:fbsde-b}
        \end{empheq} \label{eq:fbsde}%
    \end{subequations}
    where the functions $f$, $G$, $h$, and ${\varphi}$ satisfy the regularity conditions:
    \begin{itemize}
        \item $f$, $G$, $h$, and ${\varphi}$ are continuous,
        \item $f$ and $G$ are uniformly Lipschitz in $x$,
        \item $h$ satisfies quadratic growth condition in $z$,
    \end{itemize}
    such that \Cref{eq:fbsde} exists a pair of unique strong solutions \citep{yong1999stochastic,kobylanski2000backward}.
    Then, the solution to \Cref{eq:fbsde} coincides with the solution to the following second-order parabolic PDE $v(x,t) \in C^{2,1}$
    \begin{equation}
        \label{eq:nfk-pde}
        \begin{split}
        &\partial_t v + \frac{1}{2}\Tr(\nabla^2 v~GG^\T) + \nabla v^\T f + h(t,x,v,G^\T \nabla v) = 0, \quad v_1(x) = \varphi(x),
    \end{split}
    \end{equation}
    in the sense that $Y_t$ equals almost surely to $v$ along the paths generated by \Cref{eq:fbsde-f},
    and the following stochastic representation holds
    \begin{align}
        \label{eq:nfk-tf}
        v(X_t, t) = Y_t, \quad G(X_t,t)^\T \nabla v(X_t,t) = Z_t.
    \end{align}
    Intuitively, since $v$ appearing in \Cref{eq:nfk-tf} takes the random vector $X_t$ as
    its argument, $Y_t \coloneqq v(X_t, t)$ shall also be understood as a random variable.
    The dynamics of this random variable $Y_t$ is characterized by the SDE in \Cref{eq:fbsde-b}.
    One can check that, \eg the processes $(X_t, Y_t)$ in \Cref{thm:1} can be recovered from \Cref{eq:fbsde,eq:nfk-pde} via
    \begin{align*}
        f := \sigma^2\nabla\log\Psi,\quad
        G := \sigma,\quad
        h := -\frac{1}{2}\norm{Z_t}^2 - V,\quad
        v := \log\Psi.
    \end{align*}
    
\end{itemize}

\subsection{Main Proofs}
\noindent {\bf Proof of \Cref{thm:1}}$\quad$
    While Theorem~1 is a direct consequence of the nonlinear Feynman-Kac lemma, here we provide an alternative derivation using standard stochastic calculus.
    Recall that the optimal SDE on the time coordinate $t$ in \Cref{eq:sde}, restated below.
    \begin{align}
        \rd X_t &= \sigma^2\nabla \log \Psi(X_t, t) \dt + \sigma \rd W_t,~~X_0 \sim \mu,~X_1 \sim \nu \tag{\labelcref*{eq:sde}} \label{eq:sde-app}
    \end{align}
    Apply It\^o formula in \Cref{eq:ito} with $v \coloneqq \log \Psi(X_t,t)$, where $X_t$ follows \Cref{eq:sde-app},
    \begin{align*}
        \rd \log \Psi
        &= \fracpartial{\log \Psi}{t}\dt
        + \br{\nabla \log \Psi \cdot \sigma^2 \nabla \log \Psi + \frac{\sigma^2}{2} \Delta \log \Psi } \dt
        + \sigma\nabla \log \Psi \cdot \rd W_t \\
        &= \br{\fracpartial{\log \Psi}{t} +  \norm{\sigma \nabla \log \Psi}^2 + \frac{\sigma^2}{2} \Delta \log \Psi } \dt
        + \sigma\nabla \log \Psi \cdot \rd W_t,
        \label{eq:logpsi-sde} \numberthis
    \end{align*}
    and notice that the PDE $\fracpartial{\log \Psi}{t}$ obeys
    \begin{align}
        \fracpartial{\log \Psi}{t}
        = {\frac{1}{\Psi}} \pr{
         -\frac{\sigma^2}{2} \Delta \Psi + V \Psi
        }
        =  -\frac{1}{2} \norm{\sigma\nabla \log \Psi}^2 - \frac{\sigma^2}{2} \Delta\log \Psi + V,
        \label{eq:logpsi-pde}
    \end{align}
    where the second equality follows from \Cref{lm:1}.
    Substituting \Cref{eq:logpsi-pde} to \Cref{eq:logpsi-sde} yields
    \begin{align}
    \rd \log \Psi
    &= \br{
        \frac{1}{2}\norm{\sigma \nabla \log \Psi}^2 + V} \dt + \sigma \nabla \log \Psi \cdot \rd W_t. \label{eq:thm1-Y}
    \end{align}
    Now, apply the same It\^o formula by instead substituting $v \coloneqq \log \Psihat(X_t,t)$, where $X_t$ again follows \Cref{eq:sde-app},
    \begin{align}
        \rd \log \Psihat
        &= \fracpartial{\log \Psihat}{t}\dt
        + \br{\nabla \log \Psihat \cdot \sigma^2 \nabla \log \Psi + \frac{\sigma^2}{2} \Delta \log \Psihat } \dt
        + \sigma\nabla \log \Psihat \cdot \rd W_t,
        \label{eq:logpsihat-sde}
    \end{align}
    and notice that the PDE of $\fracpartial{\log \Psihat}{t}$ obeys
    \begin{align}
        \fracpartial{\log \Psihat}{t}
        = {\frac{1}{\Psihat}} \pr{
         \frac{\sigma^2}{2} \Delta \Psihat - V \Psihat
        }
        =  \frac{1}{2} \norm{\sigma\nabla \log \Psihat}^2 + \frac{\sigma^2}{2} \Delta\log \Psihat - V,
        \label{eq:logpsihat-pde}
    \end{align}
    where the second equality again follows by \Cref{lm:1}.
    Substituting \Cref{eq:logpsihat-pde} to \Cref{eq:logpsihat-sde} yields
    \begin{align*}
        \rd \log \Psihat
        &= \br{
        \frac{1}{2} \norm{\sigma\nabla \log \Psihat}^2 - V + \sigma^2\Delta\log \Psihat + \sigma^2\nabla \log \Psihat \cdot \nabla \log {\Psi}
        } \dt + \sigma \nabla \log \Psihat \cdot \rd W_t \\
        &= \br{
            \frac{1}{2} \norm{\sigma \nabla \log \Psihat}^2 - V
            + \nabla \cdot (\sigma^2 \nabla \log \Psihat)
            + \sigma^2\nabla \log \Psihat \cdot \nabla \log {\Psi}
        } \dt + \sigma \nabla \log \Psihat \cdot \rd W_t.
        \label{eq:thm1-Yhat} \numberthis
    \end{align*}
    We conclude the proof by substituting
    $Y_t \coloneqq \log \Psi(X_t,t)$ and $\Yhat_t \coloneqq \log \Psihat(X_t,t)$
    into \Cref{eq:thm1-Y,eq:thm1-Yhat}. \hfill $\blacksquare$ 

\vspace{0.3in}

\noindent {\bf Proof of \Cref{thm:2}}$\quad$
We restrict our focus on $\ell_\mathrm{fwd}(\theta, \phi)$, as the derivation of $\ell_\mathrm{bwd}(\phi, \theta)$ follows similarly.
Let $p^\theta$ and $p^\phi$ denote the path densities induced respectively by the Neural SDEs in \Cref{eq:sdeq1,eq:sdeq2}, restated below.
\begin{align}
    \rd X_t^\theta &= \sigma Z^\theta_t(X_t^\theta) \dt + \sigma \rd W_t,\quad~ X_0 \sim \mu \tag{\labelcref*{eq:sdeq1}} \label{eq:sdeq1-app} \\
    \rd \Xbar_s^\phi &= \sigma \Zhat^\phi_s(\Xbar_s^\phi) \ds + \sigma \rd W_s,\quad \Xbar_0 \sim \nu\tag{\labelcref*{eq:sdeq2}} \label{eq:sdeq2-app}
\end{align}

We will invoke Girsanov's Theorem \citep{sarkka2019applied} to compute the KL divergence between the two path densities, $\KL(p^\theta || p^\phi)$.
Since the theorem necessitates both stochastic processes to be aligned along the \emph{same} time coordinate, we rewrite the stochastic process in \Cref{eq:sdeq1-app} with its \emph{reversed} form, following standard practices in, \eg \citet{anderson1982reverse}
\begin{align}
    \rd X_t^\theta &= \br{\sigma Z^\theta_t(X_t^\theta) - \sigma^2 \nabla \log p_t^\theta(X_t^\theta) } \dt + \sigma \rd \bar W_t,\quad~ X_1^\theta \sim p^\theta_1, \label{eq:sdeq11}
\end{align}
where $\bar W_t$ is a backward Wiener process along $t$
and $p^\theta_t$ denote the marginal density of \Cref{eq:sdeq1-app} at time $t$.
Note that \Cref{eq:sdeq11} is equivalent to \Cref{eq:sdeq1-app}---in the sense that they yield the same joint distribution and therefore time marginal---except evolving backward in $t$, or, equivalently, forward in $s$.
With that, the KL divergence can be computed as follows
\begin{align*}
    \KL(p^\theta||p^\phi)
    &= \KL(p_1^\theta||\nu) + \int_0^1 \E_{p^\theta_t}
    \br{
        \frac{1}{2}\norm{Z^\theta_t - \sigma \nabla \log p_t^\theta + \Zhat^\phi_t }^2
    } \dt,   \numberthis \label{eq:kl}
\end{align*}
where the decomposition follows by the chain rule of KL divergence.

Next, we derive the model likelihood $\log \mu^\theta$.
Notice that the marginal density $p_t^\theta(x)$ obeys the Fokker Plank equation
\begin{align*}
    \fracpartial{p_t^\theta(x)}{t}
    &= - \nabla \cdot \Big( {  p_t^\theta(x) {  \sigma Z^\theta_t(x) }  } \Big) + \frac{\sigma^2}{2} \Delta p_t^\theta(x) \numberthis \label{eq:fpe}  \\
\Rightarrow    \fracpartial{\log p_t^\theta}{t} &= \frac{1}{p_t^\theta} \Big(
        - p_t^\theta~\nabla \cdot \pr{  \sigma Z^\theta_t }
        - \nabla p_t^\theta \cdot \sigma Z^\theta_t
        + \frac{\sigma^2}{2} \Delta p_t^\theta
    \Big) %
    \\
    &= - \nabla \cdot \pr{  \sigma Z^\theta_t } - {\nabla \log p_t^\theta} \cdot {  \sigma Z^\theta_t } + \frac{\sigma^2}{2} \Delta \log p_t^\theta + \frac{1}{2}\norm{\sigma \nabla \log p_t^\theta}^2,
    \label{eq:partial_logp} \numberthis
\end{align*}
where the last equality follows by \Cref{lm:1}.
Now, invoke It\^o formula
\begin{align*}
    \rd \log p_t^\theta &=
    \fracpartial{\log p_t^\theta}{t} \dt + \br{
        {\nabla \log p_t^\theta} \cdot {  \sigma Z^\theta_t} + \frac{\sigma^2}{2} \Delta \log p_t^\theta
    } \dt + \sigma {\nabla \log p_t^\theta} \cdot \rd W_t \\
    &=
    \br{
        - \nabla \cdot \pr{  \sigma Z^\theta_t } + \sigma^2 \Delta \log p_t^\theta
        + \frac{1}{2} \norm{\sigma\nabla \log p_t^\theta}^2
    } \dt + \sigma {\nabla \log p_t^\theta} \cdot \rd W_t,
    \numberthis \label{eq:dlogp}
\end{align*}
where the last equality follows by \Cref{eq:partial_logp}.
\Cref{eq:dlogp} facilitates a computation of the parametrized model likelihood
\begin{align*}
    \log \mu^\theta(X_0) &= \E_{p_1^\theta}[\log p_1^\theta(X_1^\theta)] - \int_0^1 \E_{p_t^\theta} \br{\rd \log p_t^\theta}  \\
     &=
    \E_{p_1^\theta}[\log p_1^\theta(X_1^\theta)] - \int_0^1 \E_{p_t^\theta} \br{
        - \nabla \cdot \pr{  \sigma Z^\theta_t } + \sigma^2 \Delta \log p_t^\theta
        + \frac{1}{2} \norm{\sigma\nabla \log p_t^\theta}^2
    } \dt \\
    &=
    \E_{p_1^\theta}[\log p_1^\theta(X_1^\theta)] - \int_0^1 \E_{p_t^\theta} \br{
        - \nabla \cdot \pr{  \sigma Z^\theta_t }
        - \frac{1}{2} \norm{\sigma\nabla \log p_t^\theta}^2
    } \dt, \numberthis \label{eq:nll-app}
\end{align*}
where the last equality follows by \Cref{eq:int-by-part}.
Although \Cref{eq:nll-app} implies that the parametrized model likelihood is intractable, due to the unknown score function $\nabla \log p_t^\theta$,
combining this likelihood with the KL divergence in \Cref{eq:kl} yields a tractable objective:
\begin{align*}
    &\quad~\KL(p^\theta || p^\phi) - \log \mu^\theta(X_0) \\
    &=\KL(p_1^\theta || \nu) + \int_0^1 \E_{p^\theta_t} \br{
        \frac{1}{2}\norm{Z^\theta_t - \sigma \nabla \log p_t^\theta + \Zhat^\phi_t }^2
    } \dt \\
    &\qquad- \E_{p_1^\theta}[\log p_1^\theta(X_1^\theta)] + \int_0^1 \E_{p_t^\theta} \br{
        - \nabla \cdot \pr{  \sigma Z^\theta_t }
        - \frac{1}{2} \norm{\sigma\nabla \log p_t^\theta}^2
    } \dt \\
    &= -\E_{p_1^\theta}[\log \nu(X_1^\theta)] + \int_0^1 \E_{p^\theta_t} \br{
        \frac{1}{2}\norm{Z^\theta_t - \sigma \nabla \log p_t^\theta + \Zhat^\phi_t }^2
        - \nabla \cdot \pr{  \sigma Z^\theta_t }
        - \frac{1}{2} \norm{\sigma\nabla \log p_t^\theta}^2
    } \dt \\
    &= -\E_{p_1^\theta}[\log \nu(X_1^\theta)] + \int_0^1 \E_{p^\theta_t} \br{
        \frac{1}{2}\norm{Z^\theta_t + \Zhat^\phi_t }^2
        - \nabla \cdot \pr{  \sigma Z^\theta_t }
        - \pr{ Z^\theta_t + \Zhat^\phi_t } \cdot \sigma\nabla \log p_t^\theta
    } \dt \\
    &= -\E_{p_1^\theta}[\log \nu(X_1^\theta)] + \int_0^1 \E_{p^\theta_t} \br{
        \frac{1}{2}\norm{Z^\theta_t + \Zhat^\phi_t }^2
        - \nabla \cdot \pr{  \sigma Z^\theta_t }
        + \nabla \cdot \pr{  \sigma Z^\theta_t + \sigma \Zhat^\phi_t }
    } \dt \qquad\qquad\qquad\qquad\qquad\qquad   \text{ \color{gray} by \Cref{eq:int-by-part2}}  \\
    &= -\E_{p_1^\theta}[\log \nu(X_1^\theta)] + \int_0^1 \E_{p^\theta_t} \br{
        \frac{1}{2}\norm{Z^\theta_t + \Zhat^\phi_t }^2
        + \nabla \cdot \pr{  \sigma \Zhat^\phi_t }
    } \dt \\
    &= \ell_\mathrm{fwd}(\theta, \phi).
\end{align*}
Next, we proceed to the minimizer of $\ell_\mathrm{fwd}(\theta, \phi)$.
Since $\phi$ appears in $\ell_\mathrm{fwd}(\theta, \phi)$ only through the KL divergence in $\KL(p^\theta || p^\phi)$,
$\phii$ must satisfy the following equality \citep{nelson2020dynamical}
\begin{align}
    \Zhat^\phii_t \coloneqq \sigma \nabla \log p_t^\theta - Z^\theta_t.
    \label{eq:nelson}
\end{align}
The remaining objective is thereby
\begin{align*}
    \ell_\mathrm{fwd}(\theta, \phii)
    &= \E_{p_1^\theta}[-\log \nu(X_1^\theta)] + \int_0^1 \E_{p^\theta_t} \br{
        - \nabla \cdot \pr{  \sigma Z^\theta_t }
        - \frac{1}{2}\norm{\sigma \nabla \log p_t^\theta }^2
    } \dt \\
    &= \E_{p_1^\theta}[-\log \nu(X_1^\theta)] + \int_0^1 \E_{p^\theta_t} \br{
        - \nabla \cdot \pr{  \sigma Z^\theta_t }
        + \nabla \cdot \pr{  \frac{\sigma}{2} Z^\theta_t + \frac{\sigma}{2} \Zhat^\phii_t }
    } \dt \\
    &= \E_{p_1^\theta}[-\log \nu(X_1^\theta)] - \int_0^1 \E_{p^\theta_t} \br{
        \nabla \cdot \pr{  \frac{\sigma}{2} \pr{ Z^\theta_t - \Zhat^\phii_t} }
    } \dt, \label{eq:a} \numberthis
\end{align*}
where the second equality follows by \Cref{eq:int-by-part2,eq:nelson}.
Importantly, \Cref{eq:a} is exactly the likelihood objective for a neural ordinary differential equations (ODE) \citep{grathwohl2019ffjord} with a parameterized vector field $\frac{\sigma}{2} ({ Z^\theta_t - \Zhat^\phii_t})$.
Indeed, the equivalent ODE representation of \Cref{eq:sdeq1-app} is known as
 \begin{align*}
     \rd X_t^\theta
     = \frac{\sigma}{2} \br{ Z^\theta_t(X_t^\theta) - \Zhat^\phii_t(X_t^\theta) } \dt,
\end{align*}
which can be shown by rewriting the Fokker-Plank equation in \Cref{eq:fpe} as
\begin{align*}
    \fracpartial{p_t^\theta(x)}{t}
    &= - \nabla \cdot \Big( {  p_t^\theta(x) {  \sigma Z^\theta_t(x) }  } \Big) + \frac{\sigma^2}{2} \Delta p_t^\theta(x) \\
    &= - \nabla \cdot \Big( {  p_t^\theta(x) {  \sigma Z^\theta_t(x) } - \frac{\sigma^2}{2} p_t^\theta(x) \nabla \log p_t^\theta(x)  } \Big) \\
    &= - \nabla \cdot \Big( {  p_t^\theta(x) \Big( \frac{\sigma}{2} Z^\theta_t(X_t^\theta) - \frac{\sigma}{2}\Zhat^\phii_t(X_t^\theta) \Big) } \Big),
\end{align*}
where the last equality follows by \Cref{eq:nelson}.
Consequently, the minimizer $\thetaa$ achieves
\begin{align}
    \ell_\mathrm{fwd}(\thetaa, \phii) = -\log \mu,
    \label{eq:nll-min}
\end{align}
provided the universal approximation theorem of neural networks \citep{hornik1989multilayer,csaji2001approximation}.
We conclude the proof. \hfill $\blacksquare$ 

\vspace{0.3in}

\noindent {\bf Proof of \Cref{thm:3}}$\quad$
Suppose the parametrized functions,
\begin{align*}
    Y^\theta_t(x) \equiv Y^\theta(x,t),\quad
    \Yhat^\phi_t(x) \equiv \Yhat^\phi(x,t), \quad
    Z_t^\theta(x) \coloneqq \sigma \nabla Y^\theta(x,t), \quad
    \Zhat_t^\phi(x) \coloneqq \sigma \nabla \Yhat^\phi(x,t)
\end{align*}
satisfy the following set of SDEs
\begin{subequations}
\begin{align}
    \rd X_t^\theta &= \sigma Z^\theta_t(X_t^\theta) \dt + \sigma \rd W_t \label{eq:x} \\
    \rd Y_t^\theta &= \pr{\frac{1}{2} \norm{Z^\theta_t(X_t^\theta)}^2 + V(X_t^\theta,t) } \dt + Z^\theta_t(X_t^\theta) \cdot \rd W_t \label{eq:y-app} \\
    \rd \Yhat_t^\phi &= \pr{
        \frac{1}{2} \norm{\Zhat_t^\phi(X_t^\theta)}^2
        - V(X_t^\theta,t)
        + \nabla \cdot (\sigma \Zhat_t^\phi(X_t^\theta))
        + Z_t^\theta(X_t^\theta) \cdot \Zhat_t^\phi(X_t^\theta)
    } \dt + \Zhat_t^\phi(X_t^\theta) \cdot \rd W_t \label{eq:yhat-app}
\end{align}
\end{subequations}
with the boundary conditions
\begin{align*}
    X_0 \sim \mu, \quad
    X_1^\theta \sim  p_1^\theta \equiv \nu, \quad
    Y_0^\theta + \Yhat_0^\phi = \log \mu, \quad
    Y_1^\theta + \Yhat_1^\phi = \log \nu.
\end{align*}
Then, it can be readily seen that \Cref{eq:yhat-app} minimizes $\ell_\mathrm{TD}(\phi)$ by construction.
On the other hand, \Cref{eq:y-app,eq:yhat-app} suggest that
\begin{align*}
    - \pr{Y_0^\theta + \Yhat_0^\phi}
    &=  - \E_{p_1^\theta} \br{ Y_1^\theta + \Yhat_1^\phi }
        + \int_0^1 \E_{p_t^\theta}\br{
        \rd Y_t^\theta + \rd \Yhat_t^\theta
    } \\
    \Rightarrow - \log \mu(X_0) &=
        - \E_{p_1^\theta} \br{\log \nu(X_1^\theta)}
        + \int_0^1 \E_{p^\theta_t} \br{
        \frac{1}{2}\norm{Z^\theta_t + \Zhat^\phi_t }^2
        + \nabla \cdot \pr{  \sigma \Zhat^\phi_t }
    } \dt, %
\end{align*}
implying $\ell_\mathrm{fwd}(\theta,\phi)$ is also minimized due to \Cref{eq:nll-min}.
This concludes the necessary condition.
For the sufficient condition, 
we wish to show that $Y^\thetaa_t$, $\Yhat^\phii_t$, $Z_t^\thetaa$, and $\Zhat_t^\phii$ satisfy \Cref{eq:y-app,eq:yhat-app}
given $\thetaa, \phii$ that minimize $\ell_\mathrm{fwd}(\theta,\phi)$ and $\ell_\mathrm{TD}(\phi)$.
Since $\ell_\mathrm{TD}(\phii) = 0$ readily implies \Cref{eq:yhat-app},
we focus primarily on showing \Cref{eq:y-app}.
More precisely, the question pertains to showing that $Y^\thetaa_t$ must obey \Cref{eq:y-app} if $\thetaa$ solves
\begin{align}
    -\log \mu(X_0) &= \min_{\theta} \ell_\mathrm{fwd}(\theta,\phii) \nonumber \\
    &=
    \min_\theta \Big\{
        -\E_{p_1^\theta}[\log \nu(X_1^\theta)] + \int_0^1 \E_{p^\theta_t} \br{
        \frac{1}{2}\norm{Z^\theta_t + \Zhat^\phii_t }^2
        + \nabla \cdot \pr{  \sigma \Zhat^\phii_t }
    } \dt
    \Big\}. \label{eq:dev}
\end{align}
Now, notice that $\phii$ satisfies
\begin{align}
    \E_{p_1^\theta}\br{ \Yhat_1^\phii} -  \Yhat_0^\phii =
    \int_0^1 \E_{p_t^\theta} \br{
        \frac{1}{2} \norm{\Zhat_t^\phii}^2
        - V
        + \nabla \cdot (\sigma \Zhat_t^\phii)
        + Z_t^\theta \cdot \Zhat_t^\phii
    } \dt.
    \label{eq:dev2}
\end{align}
Substituting \Cref{eq:dev2} into \Cref{eq:dev} yields
\begin{align*}
    -\log \mu(X_0) &=
    \min_{\theta} \Big\{
        \E_{p_1^\theta}\Big[ -\log \nu(X_1^\theta) + \Yhat_1^\phii \Big]
        - \Yhat_0^\phii
        + \int_0^1 \E_{p_t^\theta} \br{
            \frac{1}{2} \norm{Z^\theta_t}^2 + V
        } \dt
    \Big\} \\
    -\log \mu(X_0) + \Yhat_0^\phii &=
    \min_{\theta} \Big\{
        \E_{p_1^\theta}\Big[ -\log \nu(X_1^\theta) + \Yhat_1^\phii \Big]
        + \int_0^1 \E_{p_t^\theta} \br{
            \frac{1}{2} \norm{Z^\theta_t}^2 + V
        } \dt
    \Big\}. \numberthis \label{soc}
\end{align*}
\Cref{soc} is a standard stochastic optimal control problem \citep{exarchos2018stochastic,pereira2019neural}, whose optimality condition is known to obey 
\begin{align*}
    \rd Y_t^\thetaa = \pr{\frac{1}{2} \norm{Z^\thetaa_t}^2 + V } \dt + Z^\thetaa_t \cdot \rd W_t,
    ~~ - Y_1^\thetaa = -\log \nu + \Yhat_1^\phii,
    ~~ - Y_0^\thetaa = -\log \mu + \Yhat_0^\phii,
\end{align*}
which recovers \Cref{eq:y-app} and the boundary conditions. We conclude the proof.
\hfill $\blacksquare$

\section{Additional Clarifications}

\subsection{Optimality conditions of GSBs} \label{app:b.2}
Here, we provide a concise derivation of how the PDEs \Cref{eq:sb-pde} emerges from the variational formulation of GSBs \Cref{eq:sb-soc} as the \emph{optimality conditions}.
We refer readers to, \eg \cite{chen2015optimal} and \cite{liu2022deep} for a complete treatment.
We begin by rewriting the SDE constraint in \Cref{eq:sb-soc} with the diffusion-transport, \ie Fokker-Plank, equation
\begin{align}
    \partial_t \rho + \nabla\cdot(\sigma u\rho) - \frac{1}{2}\sigma^2\Delta \rho,
    \quad\rho_0 = \mu,\quad\rho_1 = \nu, \label{eq:fpe-appb}
\end{align}
where $\rho \equiv \rho(x,t)$ is the marginal density at time $t$, induced by the stochastic controlled process.
The Lagrangian formulation of \Cref{eq:sb-soc} thereby reads
\begin{align*}
    \calL(\rho, u, \lambda) = \int_{\R^d\times[0,1]} \br{ \frac{1}{2}\|u(x,t)\|^2 + V(x,t) } \rho(x,t) \rd x \rd t \\
    - \int_{\R^d\times[0,1]} \br{ \partial_t \rho + \nabla\cdot(\sigma u\rho) - \frac{1}{2}\sigma^2\Delta \rho } \lambda(x,t) \rd x \rd t,
\end{align*}
where $\lambda(x,t)$ is the Lagrange multiplier and $\rho(x,t)$ are constrained such that $\rho_0 = \mu$, $\rho_1 = \nu$.
Under mild conditions \citep{anderson1982reverse,yong1999stochastic}
such that $\rho(x,t)$ approaches zero at a sufficient speed as $\|x\|\rightarrow \infty$,
applying integration by parts yields another PDE that, together with \Cref{eq:fpe-appb}, constitutes
the optimality conditions to \Cref{eq:sb-soc}
\begin{align}
    &\partial_t \lambda - \frac{1}{2}\sigma^2\|\nabla \lambda\| + \frac{1}{2}\sigma^2\Delta \lambda + V = 0. \label{eq:hjb}
\end{align}
The optimal control to \Cref{eq:sb-soc} is obtained by $u^\star(x,t) = - \sigma \nabla\lambda(x,t)$.
Notice that \Cref{eq:hjb} is exactly the HJB equation, except its boundary condition is defined \emph{implicitly} through \Cref{eq:fpe-appb}, which are constrained by $\mu$ and $\nu$ at the boundaries.
Hence, the PDEs in \Cref{eq:hjb,eq:fpe-appb} can be understood respectively as the conditions to an ``optimal'' ``transport'' problem.
The PDEs in \Cref{eq:sb-pde} emerge from \Cref{eq:fpe-appb,eq:hjb} through a change of variables known as the Hopf-Cole transform \citep{hopf1950partial,cole1951quasi}
\begin{align}
    \Psi(x,t) \coloneqq \exp(-\lambda(x,t)),\qquad \Psihat(x,t) \coloneqq \rho(x,t)\exp(\lambda(x,t)).
    \label{eq:hc}
\end{align}
In other words, the two sets of PDEs in \Cref{eq:fpe-appb,eq:hjb} and \Cref{eq:sb-pde} are equivalent.
The Hopf-Cole transform in \Cref{eq:hc} also suggests that $\Psi(x,t)\Psihat(x,t) = \rho(x,t)$ holds not only at the boundaries but for all $t \in [0,1]$.

Summarizing, GSBs solve the variational problems in \Cref{eq:sb-soc}, whose optimality conditions are characterized by \Cref{eq:fpe-appb,eq:hjb}, or, equivalently, the coupled PDEs in \Cref{eq:sb-pde}. Given $(\Psi, \Psihat)$ that solve \Cref{eq:sb-pde}, the optimal SDEs can be simulated via \Cref{eq:sb-sde}.

\subsection{Implementation Details}

While it is theoretically sufficient to parametrize $(Y^\theta, \Yhat^\phi)$ and then infer $Z^\theta \coloneqq \sigma \nabla Y^\theta$ and $\Zhat^\phi \coloneqq \sigma \nabla \Yhat^\phi$, we have empirically found that parametrizing $(Z^\theta, \Zhat^\phi)$ with two additional neural networks often provides extra robustness.
This choice introduces additional objectives, namely $\| Z^\theta - \sigma \nabla Y^\theta \|^2$ and $\| \Zhat^\phi - \sigma \nabla \Yhat^\phi \|^2$, to ensure the alignment of, for instance, $Y^\theta \approx \log \Psi$ and $Z^\theta \approx \sigma \nabla \log \Psi$.
In such cases, $(Z^\theta, \Zhat^\phi)$ share the same architectures as $(Y^\theta, \Yhat^\phi)$ except that the last fully-connected layer outputs vectors in $\R^d$.

The Neural SDEs are discretized using the Euler-Maruyama method for computing the TD objectives as well as during inference.
For generative modeling, we discretize the time interval into 100 steps for MNIST and CelebA, and 200 steps for Cifar10.
For crowd navigation MFGs, we consider 100 steps for GMM and 300 steps for V-neck and S-tunnel.
Regarding evaluation, the FID scores reported in \Cref{fig:gen}\emph{B} use 10,000 generated images.
Finally, we set the hyper-parameter $c$ appearing in $V_\text{obs}$ to 3,000 for GMM and 1,500 for the other two MFG problems.

\vskip 0.2in
\bibliography{paper}

\end{document}